\documentclass{article} 
\usepackage[utf8]{inputenc}
\usepackage[usenames,dvipsnames]{color}
\usepackage{iclr2019_conference,times}

\usepackage{amsmath,amsfonts,bm}

\def\eqref#1{equation~\ref{#1}}

\def\1{\bm{1}}

\DeclareMathAlphabet{\mathsfit}{\encodingdefault}{\sfdefault}{m}{sl}
\SetMathAlphabet{\mathsfit}{bold}{\encodingdefault}{\sfdefault}{bx}{n}

\usepackage{hyperref}
\usepackage{url}
\usepackage{pdfpages}
\usepackage{sidecap}
\usepackage{subcaption}
\usepackage{multirow}
\usepackage{bm}
\newcommand{\STOC}{{\sc stoc}}
\newcommand{\x}{{\mathbf{x}}}
\newcommand{\zxc}{{\z_\x^c}}
\newcommand{\zyc}{{\z_\y^c}}
\newcommand{\zys}{{\z_\y^s}}
\newcommand{\zqc}{{\z_\q^c}}
\newcommand{\zrc}{{\z_\rvar^c}}
\newcommand{\zxpc}{{\z_{\x'}^c}}
\newcommand{\zxs}{{\z_\x^s}}
\newcommand{\zxps}{{\z_{\x'}^s}}
\newcommand{\muxs}{{\bm{\mu}_\x^s}}
\newcommand{\sigmaxs}{{\bm{\sigma}_\x^s}}
\newcommand{\y}{{\mathbf{y}}}
\newcommand{\z}{z}
\newcommand{\q}{{\mathbf{q}}}
\newcommand{\rvar}{{\mathbf{r}}}
\newcommand{\simil}{{s}}
\newcommand{\CC}{\color{RawSienna}CC\color{black}}
\newcommand{\CE}{\color{RoyalBlue}CE\color{black}}
\newcommand{\PM}{\color{RoyalPurple}PM\color{black}}
\newcommand{\LF}{\color{Aquamarine}LF\color{black}}
\newcommand{\CPN}{\mathit{CPN}}

\newcommand{\LVAE}{{\mathcal{L}_\mathit{VAE}}}

\title{Open-Ended Content-Style Recombination\\Via Leakage Filtering}
\author{Karl Ridgeway$^{+*}$ \& Michael C. Mozer$^{+\dagger}$ \\ 
$^+$ Department of Computer Science, University of Colorado, Boulder\\
$^*$ Sensory, Inc.\\
$^\dagger$ presently at Google Brain, Mountain View\\
\texttt{\{karl.ridgeway,mozer\}@colorado.edu} \\
}

\iclrfinalcopy
\begin{document}

\maketitle
\lhead{}
\begin{abstract}
We consider visual domains in which a class label specifies the \emph{content} of an image, and class-irrelevant properties that differentiate instances constitute the \emph{style}. 
We present a domain-independent method that permits the \emph{open-ended} recombination of style of one image with the content of another. Open ended simply means that the method generalizes to style and content not present in the training data. The method starts by constructing a content embedding using an existing deep metric-learning technique. This trained content encoder is incorporated into a variational autoencoder (VAE), paired with a to-be-trained style encoder. The VAE reconstruction loss alone is inadequate to ensure a decomposition of the latent representation into style and content. Our method thus includes an auxiliary loss, \emph{leakage filtering}, which ensures that no style information remaining in the content representation is used for reconstruction and vice versa. We synthesize novel images by decoding the style representation obtained from one image with the content representation from another. Using this method for data-set augmentation, we obtain state-of-the-art performance on
few-shot learning tasks.

\end{abstract}

In any domain involving classification, entities are distinguished not only by class label but also by attributes orthogonal to class label. For example, if faces are classified by identity, within-class variation is
due to lighting, pose, expression, hairstyle; if masterworks of art are classified by the painter, within-class variation is due to choice of subject matter.  Following tradition \citep{tenenbaum2000}, we refer to between- and within-class variation as \emph{content} and \emph{style}, respectively. What constitutes content is defined with respect to a task. For example, in a face-recognition task, identity is the content; in an emotion-recognition task, expression is the content. There has been a wealth of research focused on decomposing content and style, with the promise that decompositions might provide insight into a domain or improve classification performance. Decompositions also allow for the synthesis of novel entities by recombining the content of one entity with the style of another. Recombinations are interesting as a creative exercise (e.g., transforming the musical composition of one artist in the style of another) or for data set augmentation.

We propose an approach to content-style decomposition and
recombination. We refer to the method as \STOC, for Style Transfer onto
Open-Ended Content. Our approach is differentiated from past work in the following ways.
First, \STOC\ can transfer to novel content. In contrast, most previous work assumes the content classes in testing are the same as those in training. 
Second, \STOC\ is general purpose and can be applied to any domain. In contrast, previous work includes approaches that leverage specific domain knowledge
   (e.g., human body pose).
Third, \STOC\ has an explicit objective, \emph{leakage filtering}, designed to isolate content and style. No such explicit objective is found in most previous work, and as a result, synthesized examples may fail to preserve content as style is varied and vice versa.
Fourth, \STOC\ requires a labeling of entities by content class, but explicit style labels are not required.  In contrast, some previous work assumes supervised training of both style and content representations.

\begin{figure}
    \begin{minipage}{3.0in}
    \includegraphics[scale=0.4]{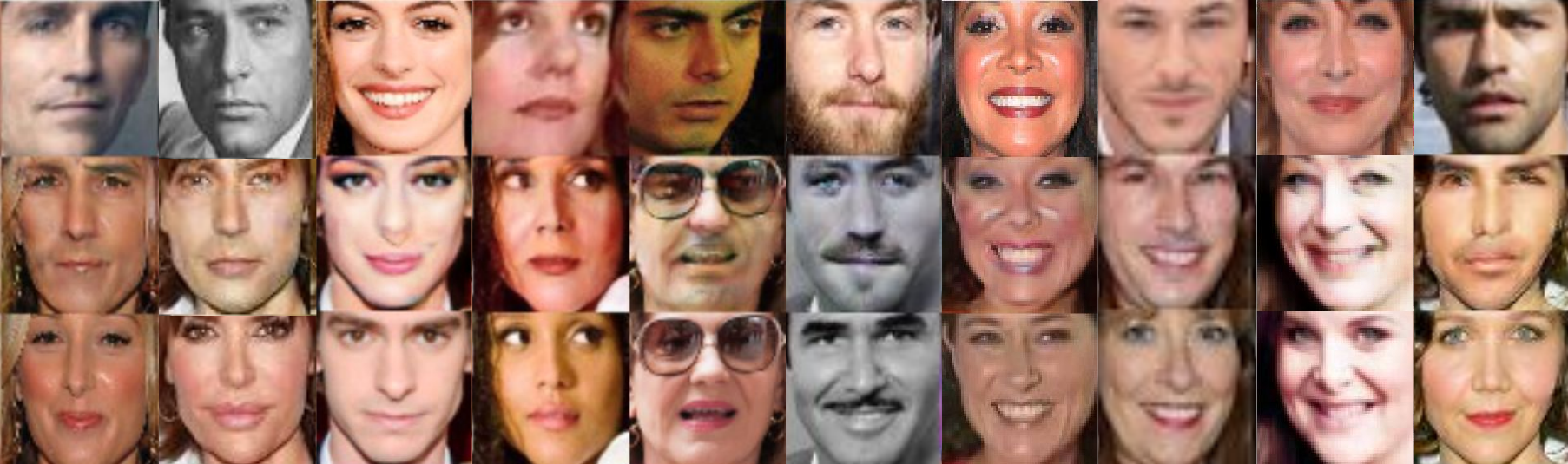}
    \end{minipage}
    \begin{minipage}{2.5in}
    \caption{Examples of content-style recombination using \STOC\ on the VGG-Face data set. The middle face in each column combines content of the top face with style of the bottom face.}
    \label{fig:faces_teaser}
    \end{minipage}
\end{figure}

Figure~\ref{fig:faces_teaser} shows examples of content-style
recombination using \STOC~on the VGG-Face~\citep{parkhi2015deep} data set.  
In each column, the content of the image in the top row is combined with the style of the
image on the bottom row to synthesize a novel image, shown in the middle row.
The images in the top and bottom rows are of identities (content) held out from the training set. Style is well maintained, and content is fairly well transferred,
at least to the degree that the faces in the middle row are more similar
to top-row than bottom-row faces. The training faces are labeled by identity, but style
is induced by the training procedure.

\section{Past research on style transfer}

A growing body of work has demonstrated impressive style transfer with models
that can translate images from one specific domain (or content class) to
another. Although some of these approaches require paired samples from both
domains \citep{pix2pix2017}, CycleGAN and its variants does not
\citep{CycleGAN2017,StarGAN2018}.  This method has been extended to exploit
constraints in video \citep{bansal2018recycle}, yielding impressive sequences
in which the mannerisms and facial movements of one individual are transferred to another.  These
methods are dependent on having \emph{many} examples from pairs of content
classes, and a model is custom trained for that pair. Therefore, the models do
not attempt to learn an explicit representation of content or to decompose
style and content.

Some domain-to-domain translation models do perform disentangling---of
the information in an entity that is shared between domains and the
information that is not shared \citep{gonzalez2018image,huang2018munit}.
\citet{huang2018munit} refer to this as content-style decomposition, but the 
range of content is quite restricted. For example, a model might be trained
to transform cats into lions, but it cannot subsequently be used to transform cats into, say, panthers.
An early proposal for style transfer \citep{kingma2014semi}, based on
variational autoencoders, can translate between more than two domains, but the model
is unable to handle novel domains in the test set.
An open-ended method such as \STOC\ can process novel content.  

Many of the above techniques are described as unsupervised because no cross-domain
correspondence between examples is required. However, from our perspective, the
separation of examples by domain \emph{is} a form of supervision, the same form
we leverage in \STOC.

Previous techniques that allow for open-ended content have typically required supervisory
signals for both content and style. That is, labels must be provided for the content class of each training sample as well as for each of a specified set of style dimensions such as pose and lighting \citep{Karaletsos2015BayesianConstraints,kulkarni2015deep,reed2014learning}.
Analogy constraints of the form $x_1:x_2 :: y_1: y_2$ have also been explored as a supervisory signal for style, specifying that two samples of one 
class $X$  have the same stylistic variation as two samples of another class, $Y$ \citep{reed2015deep}.



Methods have been developed that can transfer style to novel content without requiring explicit style labels but instead rely on domain-specific knowledge. For example, \citet{jetchev2017conditional} demonstrate the transfer of novel articles of clothing onto novel individuals, but their approach assumes that style transfer can be applied to only a masked region of the image. Other work has leveraged constraints inherent in a video sequence, either in a strong manner by extracting pose from the video \citep{brand2000style,chan2018dance,hsu2005style}, or in a weaker fashion by decomposing a video sequence into stationary (content) and nonstationary (style) components \citep{denton2017unsupervised,tulyakov2017mocogan}. 

\section{Our Approach}
Our approach builds on a Variational Autoencoder (VAE) architecture \citep{kingma2013auto}. We divide the latent code layer of the VAE into content and style components, as in the SSVAE \citep{kingma2014semi}. The content component is produced by a separately trained classifier, to be described shortly, which we will refer to as the \emph{content encoder}. The style component uses the standard VAE encoding of posterior distributions over style vectors, with a prior determined by the variational loss. It is produced by a separate network called the \emph{style encoder}. The content and (sampled) style serve as input to a \emph{decoder} net, which synthesizes an image containing the two. The VAE reconstruction loss encourages the style vector to represent any additional input variability that cannot be attributed to class (content). Content-style recombination can be achieved in the obvious manner, by synthesizing an output that is based on content of one input and style of another.

We explore four variants of this model. The baseline model, which we refer to as
\CC\ for \emph{content classifier}, uses a content encoder that is separately trained to be
a one-hot classifier using a cross-entropy loss. This model cannot handle open-ended content because the training procedure requires data from all potential content classes. Nonetheless, it is useful as a reference point for comparison to other models.  
Our second variation uses a content encoder that produces an \emph{embedding} rather than a one-hot encoding of class.  The content encoder is trained with a deep metric learning objective, the histogram loss \citep{ustinova2016learning}, which has been shown 
to have state-of-the-art performance on few-shot learning \citep{scott2018}. The embedding is $L_2$ normalized, in accordance with the fact that the histogram loss uses cosine distance. Because the content encoder produces a distributed representation of content, it can encode
novel classes and is thus in principle adequate for handling open-ended content. We call this variation of the model \CE\ for \emph{content embedding}. Both \CC\ and \CE\ use the standard VAE loss, denoted  $\LVAE$.
However, this loss does not explicitly disentangle content and style. Impurities---residual style information in the content representation and vice-versa---are problematic for content-style recombination. We thus propose two additional variations that add a \emph{decomposition loss} aimed specifically at isolating content and style: \emph{predictability minimization} (\PM), which aims to orthogonalize representations, and \emph{leakage filtering} (\LF), which aims to filter out leaks and thereby obtain better style transfer.




\subsection{Predictability Minimization}
Predictability minimization \citep{schmidhuber1992learning} encourages statistical independence between components of a representation via a loss that imposes a penalty if one component’s activation can be predicted from the others. We apply this notion to style and content
representations to minimize  content predictability from style. (Because our content encoder is frozen when training the rest of the network, we do not implement the reverse constraint.)
We build a content prediction net, or \textit{CPN}, which attempts to predict, for training sample $\x$, the output of the content encoder, $\zxc$, from the output of the style encoder, $\{ \muxs, \sigmaxs \}$. (The style encoder specifies the multivariate Gaussian style posterior obtained from the VAE.) Predictability minimization involves an adversarial loss:
\begin{equation*}
    \mathcal{L}_{\PM} = \LVAE + \lambda \min_{\theta_{\CPN}} \max_{\theta_{s}} \mathbb{E}_{\x \sim X} || \zxc  - \CPN(\muxs, \sigmaxs ) ||_2^2 ,
\end{equation*}
where $\theta_{\CPN}$ and $\theta_s$ are parameters of the CPN and style encoder, respectively, and $\lambda$ is a scaling coefficient. Training proceeds much as in a generative adversarial network \citep{goodfellow2014generative}.


\subsection{Leakage Filtering} 
One way to ensure the success of style-content recombination is to remove all style information from $\zxc$ and to remove all content information from $\zxs \sim \mathcal{N}(\muxs, \sigmaxs )$. Another way is to simply ensure that the decoder filters out any leakage of content from $\zxs$ or leakage of style from $\zxc$ in forming the reconstruction.  Leakage filtering (\LF) achieves this alternative goal via constraints that guide the training of the decoder as well as the style encoder.


\begin{figure}[b!]
    \centering
    \frame{\includegraphics[scale=0.38]{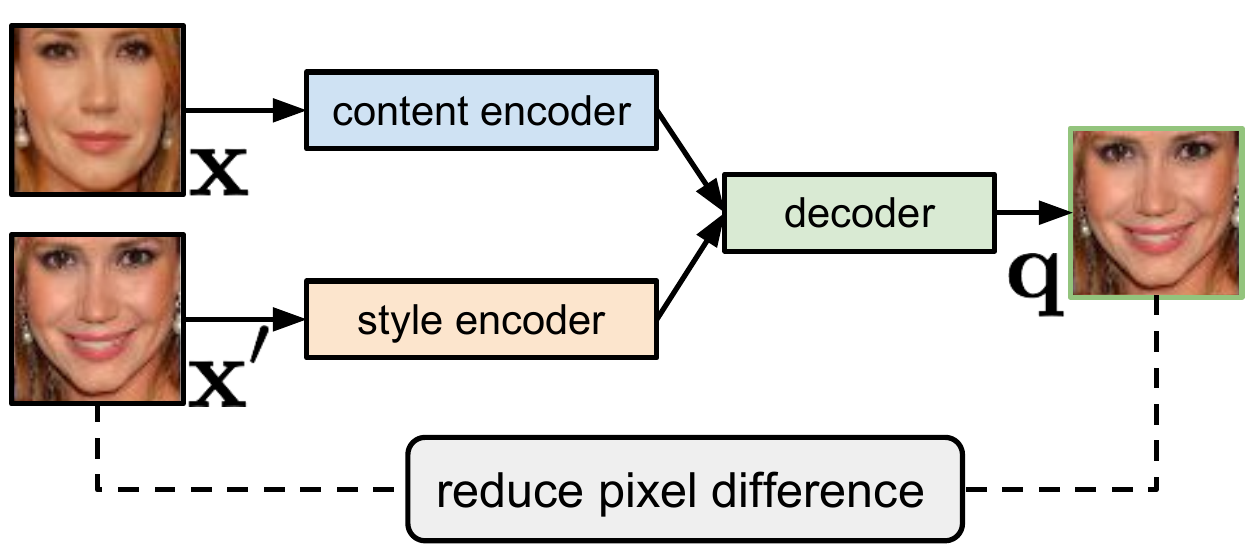}}
    \hspace{0.1in}
    \frame{\includegraphics[scale=0.38]{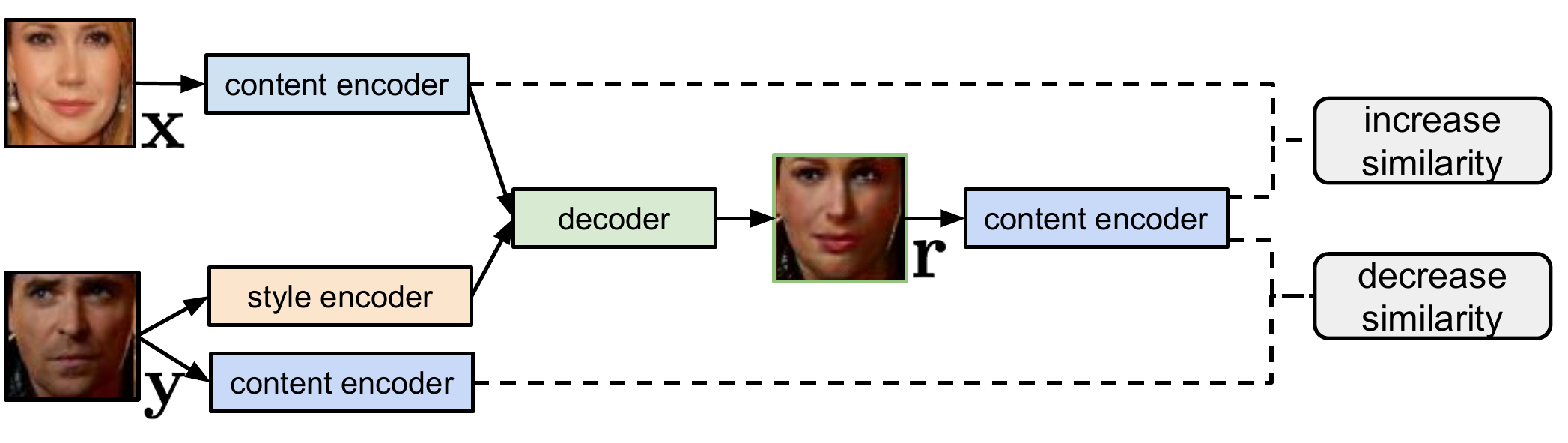}}
    \caption{
    The logic of the leakage-filtering loss. Left panel: Leakage of style from the content embedding will cause $\x'$ and $\q$ to differ. Right panel: Leakage of content from the style embedding will cause $\x$ and $\rvar$ to have dissimilar content embeddings and $\y$ and $\rvar$ to have similar embeddings.
    }
    \label{fig:leakage_filtering_illustration}
\end{figure}

The constraints of leakage filtering are illustrated in Figure~\ref{fig:leakage_filtering_illustration}.
In the left panel, we select a pair of samples of the same class, $\{\x, \x'\}$, from the complete set $P^+$, and use a decoder $D$
to recombine the style of $\x'$ with the content of $\x$ to synthesize
an image $\q$. Because $\x$ and $\x'$ have the same content class, $\q$ should be
identical to $\x'$. When they are not, style information may be leaking from $\zxc$.
In the right panel, we select a pair of samples of different classes, $\{\x, \y\}$, from the complete set $P^-$, and
transfer the style of $\y$ onto $\x$ to create a new image $\rvar$. Because $\x$ and $\rvar$ should share
the same content, the content embeddings $\zrc$ and $\zxc$ should be similar; because $\y$ and $\rvar$ do not 
share the same content, $\zrc$ and $\zyc$ should be dissimilar.  These constraints are violated when
content information leaks from the style representation, $\zys$. Just as the histogram loss was used to determine the content embedding, we repurpose the loss to quantify the similarity/dissimilarity
constraints in the content embedding. Here, however, the loss is used to adjust only parameters of decoder,
$\theta_D$, and the style encoder, $\theta_s$.

The histogram loss is based on two sets of pairwise similarity scores, $S^+$ for pairs that should be similar and
$S^-$ for pairs that should be dissimilar, as evaluated by a similarity function $\simil$; we use the cosine similarity. The histogram loss penalizes the overlap in the distributions of $S^+$ and $S^-$.
We populate $S^+$ and $S^-$ with similarities of real-to-recombined samples as well as real-to-real,
to ensure that the real-to-recombined similarities match the distributions of real-to-real:
\begin{align*}
    S^+ &= \left \{ \simil( \zxc , \zxpc ) , \simil( \zxc, \zqc ) ~|~\{\x, \x'\} \in P^+, \q = ~D( \zxc, \zxps ) \right \} ~\mathrm{and}\\
    S^- &= \left \{ \simil( \zxc, \zyc ) , \simil( \zxc, \zrc  ) ~|~\{\x, \y\} \in P^-, \rvar = ~D( \zxc , \zys ) \right \} .
\end{align*}
The histogram loss penalizes the overlap between $h^+ (.)$ and $h^- (.)$, the empirical densities formed from the sets of similarity values in $S^+$ and $S^-$, respectively. The full \LF\ loss is defined as:
\begin{equation*}
    \mathcal{L}_{\LF} = \LVAE +  \lambda_1 \left( - \mathbb{E}_{ \{\x, \x'\} \in P^+, \q = D( \zxc, \zxpc )} \log \mathrm{Pr}[ \q ~|~\x'] \right)~+~
    \lambda_2 \left( \mathbb{E}_{s\sim h^-} \left[ \int_{-\infty}^s h^+(t) dt \right] \right) ,
\end{equation*}
where $\lambda_1$ and $\lambda_2$ are scaling coefficients. Because leakage filtering imposes a cost when the decoder fails to reconstruct an image, we have found the VAE reconstruction loss to be unnecessary. In the simulations we report, we replace $\LVAE$ with $\mathcal{L}_{KL}$, the KL-divergence term of the VAE loss.

\section{Experiments with Fixed Content}
\begin{figure}[bt]
    \centering
    \begin{subfigure}[t]{0.47\linewidth}
        \centering
        \caption{Content Classifier (\CC)\vspace{-.08in}}
        \includegraphics[scale=0.27]{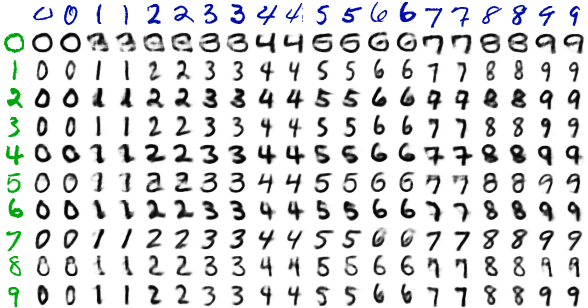}
    \end{subfigure}
    \begin{subfigure}[t]{0.47\linewidth}
        \centering
        \caption{Content Embedding (\CE)\vspace{-.08in}}
      \includegraphics[scale=0.27]{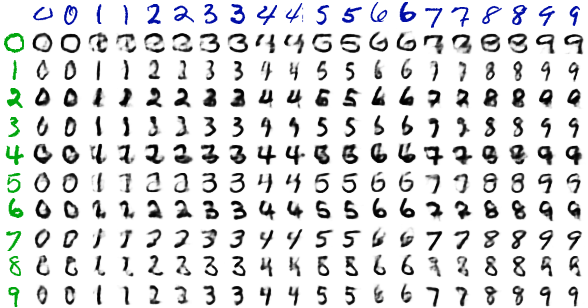}
    \end{subfigure}
    \begin{subfigure}[t]{0.47\linewidth}
        \vspace{.05in}\centering
        \caption{Predictability Minimization (\PM)\vspace{-.08in}}
        \includegraphics[scale=0.27]{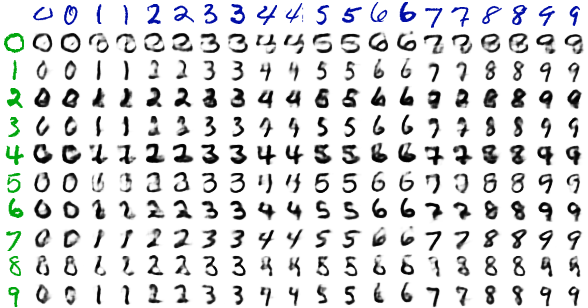}
    \end{subfigure}
    \begin{subfigure}[t]{0.47\linewidth}
       \vspace{.05in}\centering
       \caption{Leakage Filtering (\LF)\vspace{-.08in}}
       \includegraphics[scale=0.27]{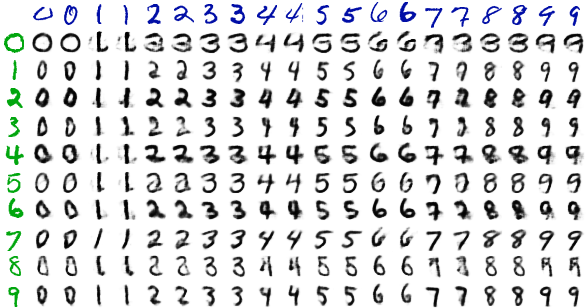}
    \end{subfigure}
    \caption{Content-style recombination on MNIST of alternative models. Black digits are synthesized from the style of green digit to the left and  content of blue digit above.}
    \label{fig:content_style_recombination_mnist}
\end{figure}

We begin with a data set having a fixed set of content classes, the MNIST handwritten digits \citep{lecun1998mnist}. Details of training, validation, and model architecture are presented in the Appendix.
%
A qualitative comparison of content-style recombination of held-out test samples for \CC, \CE, \PM, and \LF\ variations is shown
in Figure~\ref{fig:content_style_recombination_mnist}. In each case, loss weightings are hand tuned by visually
inspecting recombinations from the validation set. In general, if too much weight is placed on reconstruction,
the model will ignore content, and every row will look identical. If too much weight is placed on the decomposition
loss or KL divergence, then there will be too much uniformity in a column, with little style transfer.
In each grid of digits, the blue top row indicates
the input digit (from the test set) used to specify content. The green leftmost column indicates the input digit used
to specify style. Each gray digit is a sample from the network, with content specified by the corresponding blue digit, and
style specified by the corresponding green digit. To the extent that content-style recombination is effective,
all digits in a row should have the same style, all digits in a column should have the same class, and the two columns of each
content should be identical despite variation in the blue digits. \CC\ is superior to \CE, but this result is unsurprising:
representing content as a probability distribution over a fixed set of classes
is a stronger constraint than a content embedding. 
Variant \LF\ appears to be superior to either \PM\ or \CE, and surprisingly \LF\ appears to be as good as, or better than, \CC: the inductive bias of leakage filtering  allows it to overcome the limitations of the weaker supervisory signal of the content embedding.


For a quantitative evaluation of the quality of synthetic digits, we investigate performance of a classifier trained from scratch on synthetic digits and tested on natural digits; we call this procedure \emph{natural evaluation with synthetic training}, or \emph{NEST}. If the synthetic digits do not look natural or have little stylistic variation, test performance is poor. To synthesize digits, we first select a \emph{prototype} content representation: The prototype content embedding for a digit class is the training instance that minimizes the sum squared Euclidean
distance to all other instances of the same class. The prototype for \CC\ is simply the one-hot vector for the given class.
The classifier used for training has the same architecture as our content encoder, with 10 softmax outputs trained with a cross-entropy loss.
Training is performed on minibatches of 40 samples with randomly-selected content and style provided by a random instance in (natural digit) training set, likely of a different class. 
%

\begin{figure}[b!]
    \centering
    \begin{minipage}{6.5cm}
    \includegraphics[scale=0.65]{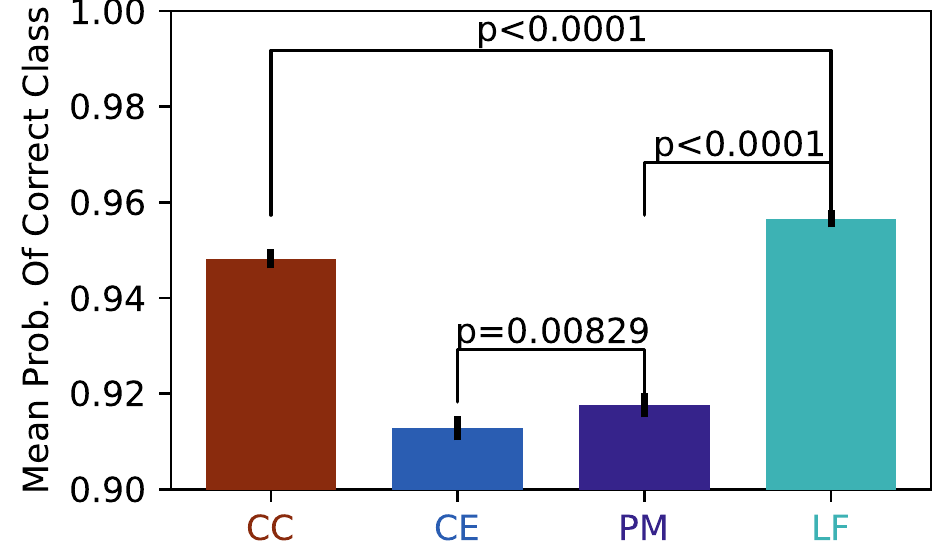}
    \end{minipage}
    \begin{minipage}{7cm}
    \caption{Naturally Evaluated, Synthetically Trained (NEST) results on MNIST. Mean probability of correct class is shown, with error bars indicating $\pm1$ standard error of the mean. $p$ values are from two-tailed Bonferroni-corrected $t$-tests with 9999 degrees of freedom. All differences are highly reliable. \CC\ = Content Classifier, \CE\ = Content Embedding, \PM\ = Predictability Minimization, \LF\ = Leakage Filtering}
    \label{fig:mnist_nest_results}
    \end{minipage}
\end{figure}

Figure~\ref{fig:mnist_nest_results} shows the mean probability of the correct class, a more sensitive metric than classification accuracy.
Both \PM\ and \LF\ outperform the baseline \CE, indicating that our losses to isolate content and style are doing the right thing.
\LF\ is clearly superior to \PM, and in fact even beats \CC, which is surprising because \LF\ allows for open-ended content whereas \CC\ does not.
Because the VAE provides a prior over style, it is possible to simply sample style from the prior, rather
than transferring it from another example. We repeated the NEST simulation using styles drawn from the prior and obtained similar results.
Having shown the superiority of \LF\ on a fixed set of classes, we next investigate performance of \LF\ with open-ended content.

\section{Experiments with Open-Ended Content}
We experiment with \LF\ on two many-class data sets: Omniglot \citep{lake2015human} and 
VGG Face \citep{parkhi2015deep}. Details of data sets and split into training, validation, and test is in the Appendix.
To improve the quality of our generated images in these more complex domains, we incorporate a 
WGAN-GP \citep{gulrajani2017improved} adversarial loss.
This additional objective requires another scaling hyperparameter for the W-GAN loss, but training is otherwise identical to the MNIST procedure.
We use a ResNet architecture for the content-style encoders, the decoder, and the critic network of the WGAN-GP.
For the VGG-Face data set,  we include U-Net \citep{ronneberger2015u} skip connections from both the style and content encoders to the decoder. Additional details can be found in the Appendix.

\vspace{-.03in}
\subsection{Qualitative Results}
\begin{figure}[t]
    \centering
    \begin{minipage}{10cm}
    \includegraphics[scale=0.3]{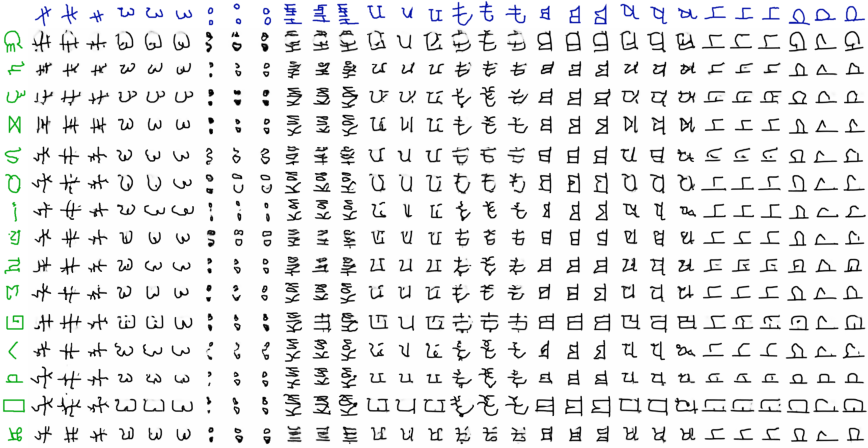}
    \end{minipage}
    \begin{minipage}{3.7cm}
    \caption{Examples of recombined content and style
    using the Omniglot dataset of handwritten characters.
    The blue characters in the top row are test samples
    used to infer
    content, and the green characters are training
    samples used to infer style. }
    \label{fig:omniglot_qualitative}
    \end{minipage}
\end{figure}

\vspace{-.06in}
Figure~\ref{fig:omniglot_qualitative} shows Omniglot characters with recombined content and style.
Content is inferred from the blue character at the top of the column, a novel class from the test set.
Style is inferred from the green character on the left, drawn from the training set. 
The content classes are repeated in order to determine how successful the model is at ignoring stylistic variation from the sample used to provide
content. The three same-class digits in a given row are not always identical, but there is certainly more variation in a column (varying style)
than there is in a row triplet (varying samples providing the content).
All characters in a row appear to share stylistic features: e.g., they are very small, have wavy lines, are bold, or are boxy in shape.

\begin{SCfigure}[][b!]
    \centering
    \includegraphics[scale=0.28]{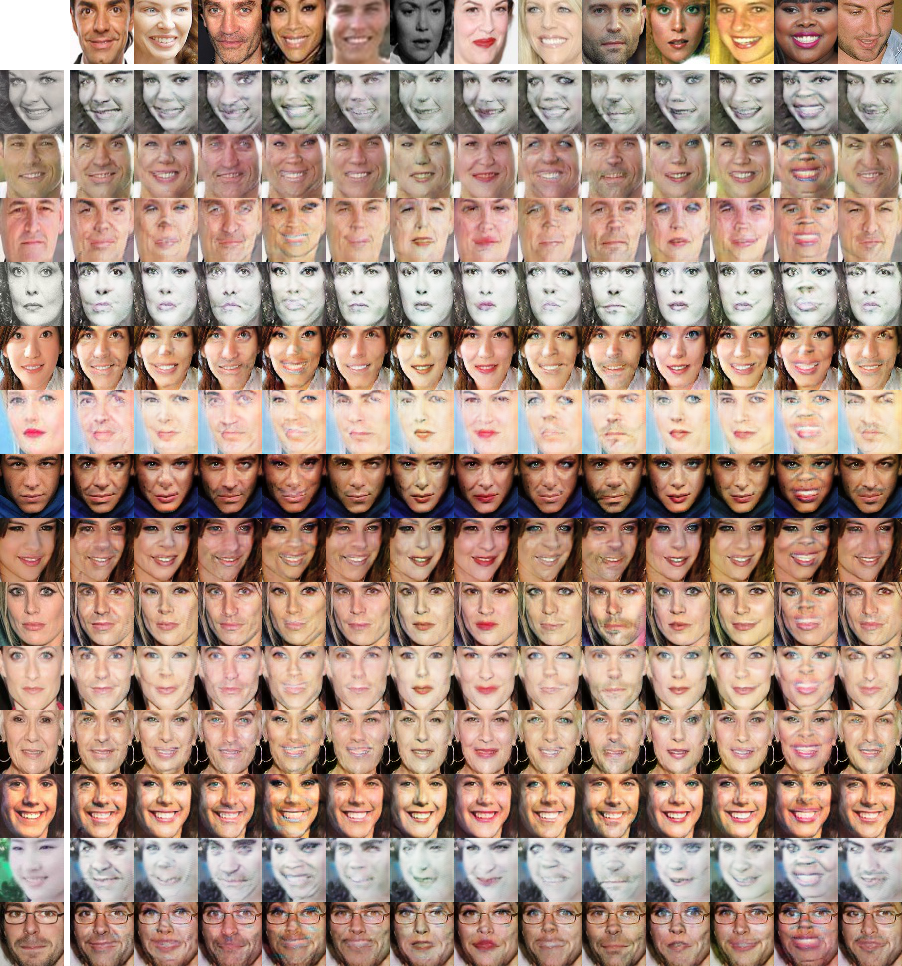}
    \caption{Examples of content-style recombination using the VGG-Face
    data set.  The images in the matrix are formed by recombining the
    content (identity) of the image in the top row with the style
    of the image in the left column.  The top row contains samples from 
    identities held out from training. The left column contains other samples
    from the data set.}
    \label{fig:vgg_qualitative}
\end{SCfigure}

Figure~\ref{fig:vgg_qualitative} shows examples of VGG Faces with recombined content (the face in the top row) and style (the face in the left column). Looking across a row,
the model preserves many aspects of style, including pose, lighting conditions, and facial expression. In the last row, even glasses are considered a stylistic feature, surprising given the strong correlation of glasses presence across instance of an individual.
Looking down a column, many identity-related features are preserved, including nose shape, eyebrow
shape, and facial structures like strong cheekbones.


\vspace{-.03in}
\subsection{Application to Data Augmentation}
\vspace{-.06in}
Next, we explore using \STOC\ for data augmentation and evaluate on few-shot learning tasks.
Data augmentation is the process of synthesizing variations of a training sample by transformations
known to preserve some attribute of interest to a task (e.g., object class), in hopes that a predictive model
will become invariant to the introduced variations.
Domain-specific techniques are very common, especially in perceptual domains, e.g., image translation and flipping.
Style transfer using \STOC\ provides a domain-agnostic method.

Recently, other researchers have used machine learning to augment data. Several methods make use of
generative adversarial nets to refine images produced by CAD programs
\citep{shrivastava2017learning,sixt2018rendergan}, but these obviously rely on significant 
domain knowledge.
\citet{devries2017dataset} generate new samples of a class 
by interpolating the hidden representations of labeled samples of that same class. 
\citet{zhu2018emotion} generates augmented faces for emotion recognition. 
Emotion is defined as the content class and a CycleGAN-like architecture is used to translate from
one emotional expression to another.
This approach works only for a fixed set of known classes and therefore cannot be directly compared to \STOC.
Two papers \citep{Rezende2016oneshot,antoniou2017data} introduce methods for generating new samples 
that share a class with a given input sample, and are shown to work with novel classes.
Only \citet{antoniou2017data} demonstrates performance on a data-augmentation task, so we choose this paper as our primary point of comparison.

\newcommand{\source}{{$\mathcal{S}$}}
\newcommand{\target}{{$\mathcal{T}$}}
\newcommand{\support}{{\ensuremath{\mathcal{T}_s}}}
\newcommand{\augmented}{{\ensuremath{\mathcal{T}_a}}}
\newcommand{\tsa}{{\ensuremath{\mathcal{T}_{sa}}}}
\newcommand{\query}{{$\mathcal{T}_q$}}
Because data augmentation should have the greatest effect in data-sparse domains, we evaluate
\STOC\ augmentation on few-shot learning, where the goal is to obtain accurate classification based on a small number of samples. Our evaluation procedure
follows \citet{scott2018}. 
The data set is divided by content-class into source (\source) and target (\target) domains. 
\source\ is split by class into a training and validation set, used to train \STOC.
We use \target\ for evaluation. Within \target,
each class has $N$ samples, which are split into $k$ \emph{support} samples (which together make up \support), 
and $(N-k)$ \emph{query} samples (\query).
Testing proceeds in \emph{episodes}, where a subset of $n$ classes is drawn from \target\ for testing. 
We then generate the augmented set (\augmented) using the content of \support, and style drawn from \source. 
A classifier is then trained using $\tsa \equiv \{ \support, \augmented \}$. Performance is reported on the classification
accuracy of \query. 
We evaluate two different methodologies. First, we compare our method to other state-of-the-art
one-shot learning methods on the Omniglot dataset. Second, we consider the case of 
training a new classifier from scratch on only \tsa.

\noindent \textbf{One-Shot Learning with Omniglot}. 
We  investigate the common one-shot Omniglot task, where the number of classes per
episode ($n$) is 20, and the number of examples per class ($k$) is 1. 
To generate \augmented, we synthesize $m$ stylistic variations of each member of \support.
We experiment with two settings, $m=0$ (no augmentation) and $m=40$.
Also, we found that limiting the variability introduced by style transfer to be important, so instead of replacing the style of the 
samples of the support set with the style of a training example, we linearly interpolate between them.

\citet{scott2018} demonstrated that the histogram-loss embedding achieves 
state-of-the-art performance on this task. We use the histogram embedding of
the content-encoder network that is trained for \STOC,
ensuring that there is no performance difference between the content embedding 
used to train the style transfer model and the embedding used for few-shot learning. 
To evaluate an episode, we first embed the \tsa\ set using the content encoder.
For each query sample, we compute its content embedding. We then compute the $L_2$ distances
between the query embedding and each embedding in \tsa. 
For each embedding in \tsa, we
assign a weight to determine the contribution strength of that sample to the overall decision.
Each real support sample is assigned a weight $w_s$, and each of the $m$ augmented samples
is assigned a weight $w_a = (1-w_s)/m$.
The probability distribution over classes is computed via a weighted softmax on the
squared distance between the query sample and the samples in $\tsa$.

For each episode, we record the average classification accuracy for all the query samples. We
run 400 episodes, each with different random subsets of test classes, and report average accuracy across the
replications. Table~\ref{tab:omniglot_few_shot_learning_results} shows the results
for our model with and without data augmentation, along with reported results
from the literature.
For this task, we find that the baseline histogram performance is already very good.
Although the improvement from data augmentation is small, it brings the histogram
embedding performance up to the level of Conv-ARC \citep{shyam2017attentive}, which is a complex, articulated, recurrent architecture with attention that performs explicit comparisons between samples.  
DAGAN~\citep{antoniou2017data} shows a bigger improvement, but it makes use of 
an auxiliary \emph{sample-selection network}, the details of which are not explained. 

\begin{table}[t!]
    \centering
    \scalebox{0.8}{
    \begin{tabular}{c|c}
        \textbf{Model Name} & \textbf{Test Accuracy} \\ \hline
        Matching Nets \citep{vinyals2016matching} & 0.938 \\
        Prototypical Networks \citep{snell2017prototypical} & 0.960 \\
        Matching Nets (DAGAN replication) \citep{antoniou2017data} & 0.969 \\
        Matching Nets + DAGAN Augmentation \citep{antoniou2017data} & 0.974 \\
        Conv. ARC \citep{shyam2017attentive} & \textbf{0.975} \\
        Histogram Embedding (our implementation) \citep{ustinova2016learning} &  0.974 \\
        \STOC\ (ours) & \textbf{0.975}
    \end{tabular}
    }
    \caption{Average query accuracy for the one-shot learning task with the Omniglot data set, with $k=5$ samples per class 
    in the support set and $n=20$ classes per episode. 
    \vspace{-.05in}}
    \label{tab:omniglot_few_shot_learning_results}
\end{table}

\noindent \textbf{Standard Classifiers with Omniglot and VGG-Face}. We also trained standard classifiers
from scratch on $\tsa$. The classifiers are convolutional nets with 4 strided convolutional
layers, followed by a ReLU activation, batch norm, and dropout with a rate of $0.5$. Each convolutional layer has a kernel size of 5
and 64 filters. To train the nets, we split \support~into training (75\%) and validation (25\%) sets, and
use the validation set to determine the number of epochs to train for.
Minibatches are composed of some mixture of real and augmented samples, and we used the validation set to determine the
ratio. 
We generate new \augmented\ augmentations for every minibatch. 
Table~\ref{tab:vanilla_classifiers} shows the results on the test samples for both Omniglot and VGG-Face data sets.
For Omniglot, we report results on the whole set of 1299 test classes, varying $k$, the number of samples per class
in the support set. In each case, the augmentation significantly improves test accuracy. 
To compare our results with DAGAN \citep{antoniou2017data},
we select a random subset of the 212 Omniglot classes, which is the size of the DAGAN test set, and we use the same test set size as DAGAN for VGG Face.
Our baseline results do not match DAGAN, which could be because
we use a different split of the data and we include novel classes in the data set formed by rotating the original data set.
However, our model shows greater improvement over its baseline for both data sets, and for every value of $k$.
Examining the synthetic faces DAGAN produces by combining content with novel style (see their Figure 5), we are not surprised that \STOC\ is
better performing:  DAGAN recombinations produce discernably different identities, in contrast to our Figure~\ref{fig:vgg_qualitative}.

\begin{table}[b]
    \centering
    \scalebox{0.8}{
        \begin{tabular}{c|c|c|c|c|c|c|c|c}
        \multicolumn{3}{c|}{}  & \multicolumn{3}{c|}{\sc \textbf{stoc}} & \multicolumn{3}{c}{\textbf{\cite{antoniou2017data}}} \\ \hline
        \multicolumn{3}{c|}{}  & \multicolumn{2}{c|}{\textit{Test Accuracy}} & & \multicolumn{2}{c|}{\textit{Test Accuracy}} & \\ \hline
        \textit{Data set} & \textit{n} & \textit{k} & \textit{Baseline} & \textit{Augmented} & \textit{Improvement} & \textit{Baseline} & \textit{Augmented} &  \textit{Improvement} \\ \hline
        \multirow{3}{*}{Omniglot} & \multirow{3}{*}{1299} & 5 & 0.261 & \textbf{0.524} & 101\% & & & \\
         &  & 10 & 0.426 & \textbf{0.639} & 50\% & & & \\
        &  & 15 & 0.543 & \textbf{0.678} & 25\% & & & \\ \hline
        \multirow{3}{*}{Omniglot}& \multirow{3}{*}{212} & 5 & 0.435 & 0.740 & \textbf{70\%} & 0.690 & 0.821 & 19\% \\  
        & & 10 & 0.571 & 0.805 & \textbf{41\%} & 0.794 & 0.862 & 9\% \\ 
        & & 15 & 0.643 & 0.784 & \textbf{22\%} & 0.820 & 0.874 & 7\% \\ \hline
        \multirow{3}{*}{VGG Face} & \multirow{3}{*}{497} & 5 & 0.087 & 0.272 & \textbf{212\%} & 0.045 & 0.126 & 180\% \\
        &  & 15 & 0.263 & 0.448 & \textbf{70\%} & 0.393 & 0.429 & 9\% \\
        &  & 25 & 0.371 & 0.504 & \textbf{36\%} & 0.580 & 0.585 & 1\% \\
    \end{tabular}

    }
    \caption{Results for training standard classifiers on un-augmented and on augmented data. The columns,
    from left to right: the tested data set, the number of classes in the training/test data sets ($n$), 
    the number of samples per class in the support set ($k$), whether or not augmentation was used,
    and test accuracy and percentage improvement over the unaugmented baseline, 
    both for our model and DAGAN \citep{antoniou2017data}.}
    \label{tab:vanilla_classifiers}
\end{table}

\section{Conclusion}
 \vspace{-.05in}
\STOC\ is effective in transferring style onto open-ended content---content that is novel with respect to the training data.
This is a challenging task: content class boundaries cannot be determined precisely in a setting where the number of potential classes is unbounded. 
As a result, it is easy for some style information to seep into a content representation. 
We introduced the leakage-filtering loss, a novel approach to isolating content and style.
Traditionally, researchers have focused on disentangling style and content: inducing representations that separate style and content into different vector components. Given the difficulty of this challenge with only content labels and no explicit labels or domain knowledge pertaining to style, we instead
focus on ensuring that the decoder, which combines style and content to reconstruct images, does not use any residual style information in the content representation or any residual content information in the style representation. Our results yield impressive visual quality and achieve
significant
boosts in performance when \STOC\ is used for augmenting data sets to train a de novo classifier.  We also explored data augmentation for few-shot
learning and achieved performance that matches state of the art, a complex highly articulated and computation intensive model. We suspect that beating 
state-of-the-art on few-shot learning is becoming increasingly difficult, given 
that 
state-of-the-art is now bumping against the ceiling on performance in the paradigm that is typically used for evaluation.


\section{Acknowledgements}
This research was supported by the National Science Foundation awards EHR-1631428 and SES-1461535. We thank Tyler Scott for providing evaluation code.

\bibliography{iclr2019_conference}
\bibliographystyle{iclr2019_conference}

\newpage
\appendix
\section{Appendix}
\subsection{Data Sets}
\textbf{MNIST}.
The MNIST database is divided into a development set consisting of 60,000 images and a test set of 10,000.
We further divide the development set into a training set of 48,000 and validation set of 12,000, which is used
to determine model hyper-parameters such as the number of epochs to train, and relative weights on the various loss functions.

\textbf{Omniglot}.
Omniglot is composed of 20 instances of 1,623 different classes of hand-written characters from 50 different alphabets. Following convention \citep{scott2018,snell2017prototypical,triantafillou2017few}, 
we augment the data set with all $90^{\circ}$ rotations, resulting in 6,492 classes.
The classes are split randomly into 4,154 training, 1,039 validation, and 1,299 test classes.

\textbf{VGG Faces}.
We use the same subset of the data as \citep{antoniou2017data},
splitting the data into 1,750 training classes, 53 validation classes, and 497 test classes.

\subsection{Network Architectures And Parameter Settings}
\begin{figure}[b]
    \centering
    \includegraphics[scale=0.4]{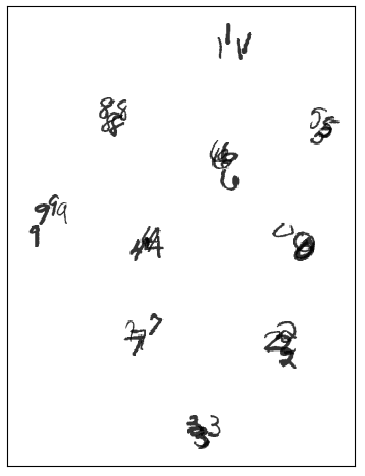}
    \hspace{0.2in}
    \caption{Example t-SNE visualization of the histogram embedding of MNIST digits. 
    }
    \label{fig:embedding_hist_tsne}
\end{figure}

\textbf{Architecture for MNIST}. The content and style encoders have the same network architecture: 
two convolutional layers with $5 \times 5$ kernels and 64 filters,
followed by a fully connected layer. Each convolutional layer includes batch normalization and leakly ReLU activation function.
The three variations with content embeddings (\CE, \PM, \LF) use the same content encoder network with 50 dimensions. 
Figure~\ref{fig:embedding_hist_tsne} shows a two-dimensional t-SNE visualization of the embedding of
the content encoder, with four randomly sampled digits of each class from the test set. Some within-class variation is preserved, but
the digit classes are still well separated.
\CC\ has a 10-dimensional (one hot) content representation. All four variations use 50 dimensions for the style representation.
The content representation is $L_2$ normalized to have a unit length,
required for the histogram loss. 
The generator network consists of a linear projection from the 100-dimensional combined content and style representation to
a $6\times 6\times 32$ tensor. This tensor is then up-sampled twice using transposed, fractionally-strided convolutions, each
with a kernel size of $5\times 5$ and 64 filters. Finally, the output of the generator is normalized to lie within $[-1,1]$
using a hyperbolic tangent activation. The inputs to the network are also normalized to lie in this range.
All networks were trained with the Adam optimizer \citep{kingma2014adam} with a learning rate of $2*10^{-4}$.
For predictability minimization, the CPN network consists of simple feed-forward net with one 100-dimensional hidden layer and a leaky ReLU activation.

\textbf{Architecture for omniglot}. 
For the content and style encoders, we start with a simple convolutional layer, followed by three ResNet blocks, with three convolutional layers
each and a skip-connection at the end. We add a final convolutional layer to reduce the output channels to 3 (RGB). The output is normalized
with a hyperbolic tangent function to lie within $[-1,1]$. The content and style representations are both 100-dimensional.
After each convolution, we use a leaky ReLU activation, and then batch re-normalization \citep{ioffe2017batch}.
Each convolutional layer has a $3 \times 3$ kernel and 48 filters. For the decoder and WGAN-GP critic, we use the an identical architecture, except
that each convolutional layer has 64 filters. As noted in \citet{gulrajani2017improved}, batch normalization violates the assumptions of WGAN-GP, so 
our critic net uses layer normalization \citep{ba2016layer}.
The model was trained for 100 epochs, using the Adam optimizer and a learning rate of $10^{-4}$.
We set the coefficient on the KL-divergence loss to $1$, and the coefficients on $\lambda_1$ and $\lambda_2$ of
$\mathcal{L}_{\LF}$ to 20, and the WGAN-GP loss is multiplied by $0.5$. We set the gradient penalty weight parameter of WGAN-GP to 10.

\textbf{Architecture for VGG-Faces}. 
We start with the same ResNet architecture that we used for omniglot, but use 64 filters in each convolutional layer of the style and content encoders.
We also include U-Net skip connections from the output of each resnet block in both the content encoder and the
style encoder to the corresponding ResNet block input in the decoder. 
The WGAP-GP critic's architecture is identical to that of the decoder, except it has no U-Net skip connections.
The content representation has 200 dimensions, and the style representation has 600.
The model was trained for 10 epochs, using the Adam optimizer and a learning rate of $10^{-4}$.
We set the coefficient on the KL-divergence loss to 1, and both of the coefficients on $\lambda_1$ and $\lambda_2$ of
$\mathcal{L}_{\LF}$ to 5, and the WGAN-GP loss is multiplied by $0.5$.
For VGG-Faces, we find a higher value of the weight on the gradient penalty parameter (100) gives better training stability.

\textbf{Minibatch Composition}. 
For MNIST, minibatches were constructed of 4 samples of each of the 10 classes, with every possible positive- and negative
comparison included in the $\mathcal{L}_{\LF}$ objective.
To train the content encoders for both omniglot and VGG-Faces, we construct minibatches by sampling 20 classes, with 10 samples each, which is close to the
recommended batch size of 256 from \citet{ustinova2016learning}. For performance reasons, we
sample only 10 classes with 3 samples each to train the style encoder, generator, and WGAN-GP critic. We sub-sample
the between-class comparisons such that their count equals the count of within-class comparisons.

\textbf{Parameters for One-Shot Learning}. 
We find that setting the weight on the real support data $w_S$ to a value around $0.85$ to give the best results. We also found
that a relatively low temperature in the softmax fuction works better than a high temperature (we used $T=0.05$).

\subsection{Exploration of Loss Coefficients On MNIST}
\begin{figure}
    \centering
    \begin{subfigure}[t]{0.45\linewidth}
        \centering
        \includegraphics[scale=0.4]{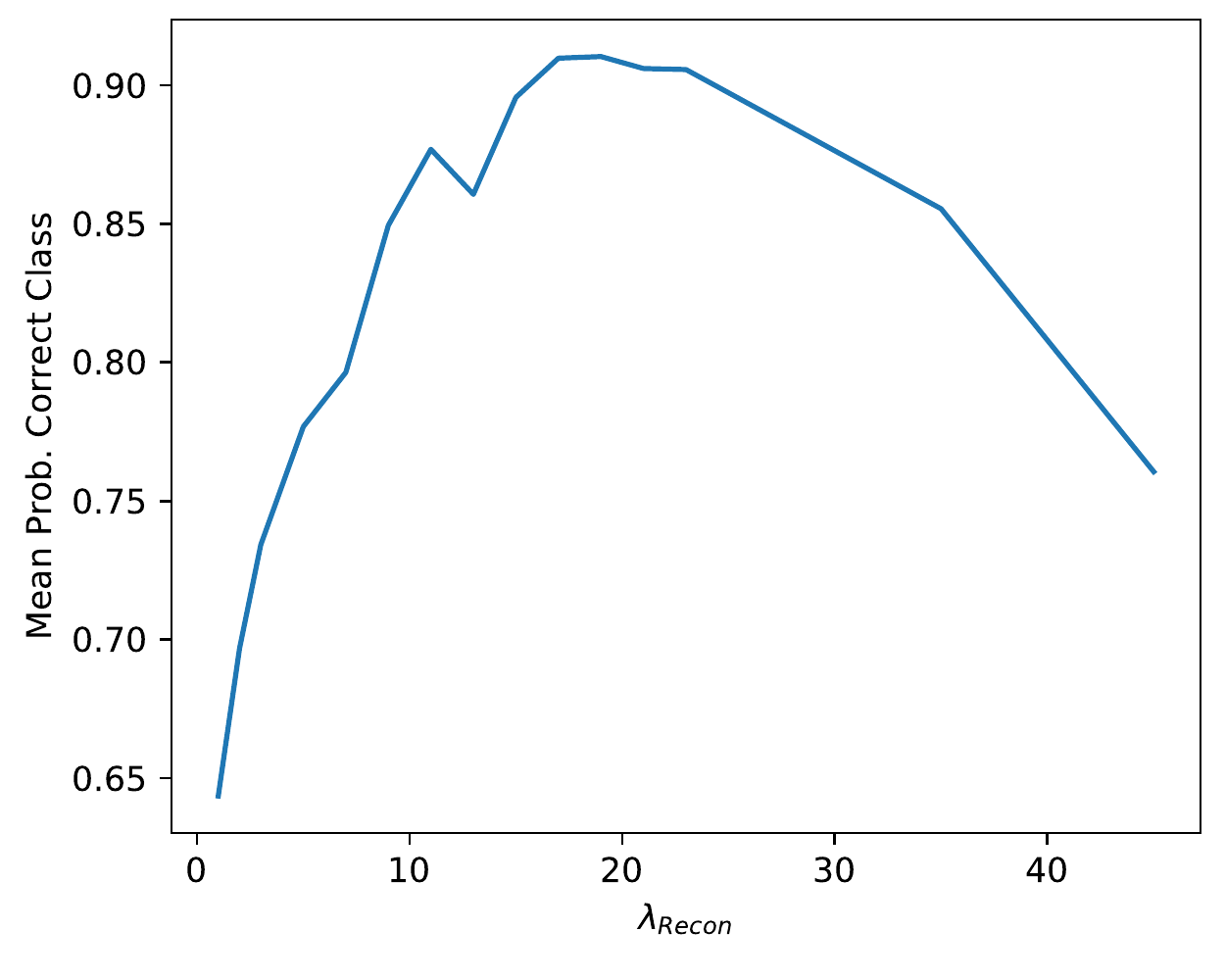}
        \caption{NEST validation performance as a function of weight on reconstruction objective for \CE.}
    \end{subfigure}
    \hspace{0.2in}
    \begin{subfigure}[t]{0.45\linewidth}
        \centering
        \includegraphics[scale=0.4]{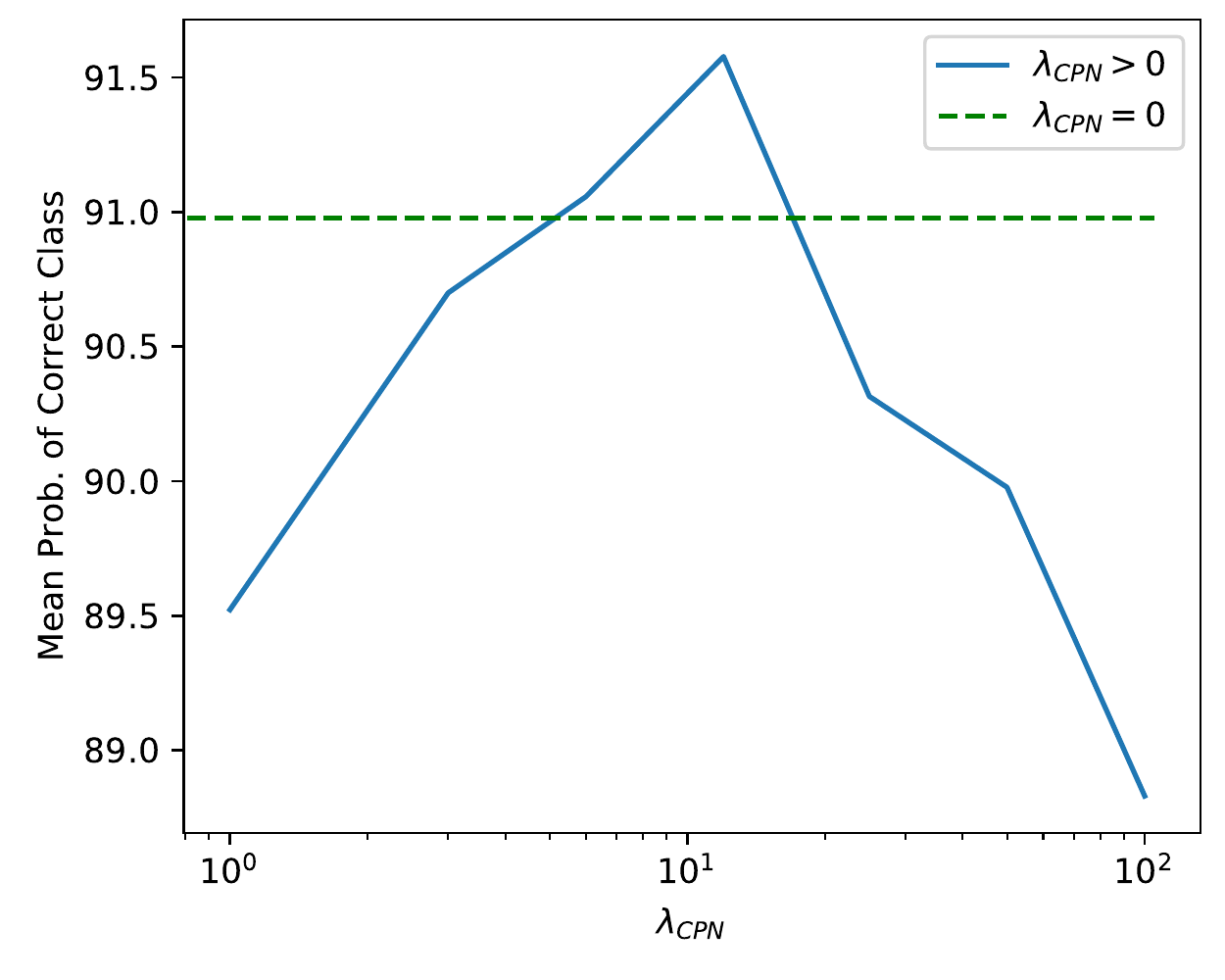}
        \caption{NEST validation performance as a function of weight on predictability minimization objective for \PM.}
    \end{subfigure}
    \begin{subfigure}[t]{\linewidth}
        \centering
        \includegraphics[scale=0.5]{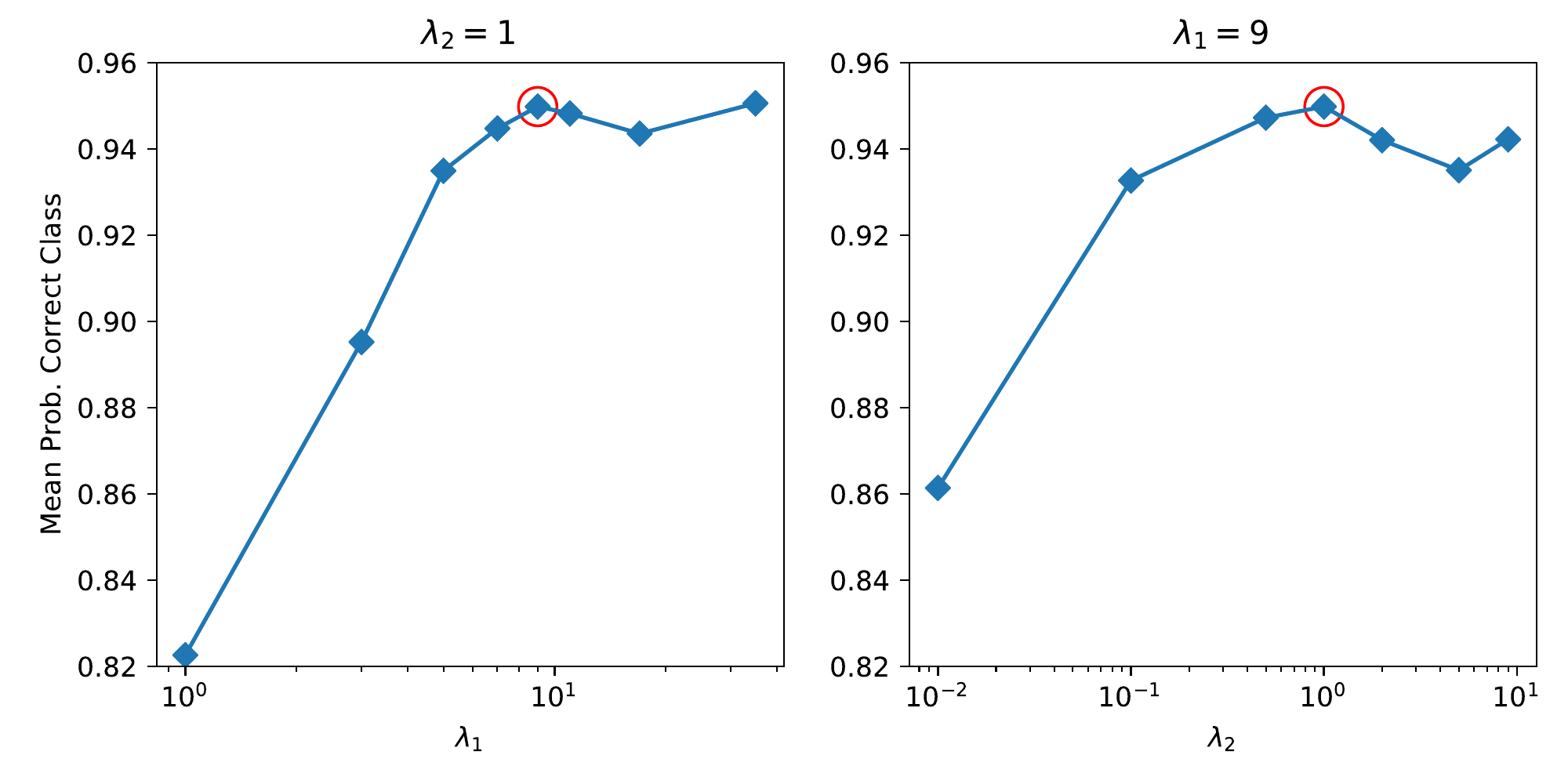}
        \caption{NEST validation performance as a function of weights on \LF. The best-performing model is highlighted in red.}
    \end{subfigure}
    \caption{Explorations of loss weights for \CE, \PM, and \LF.}
    \label{fig:loss_weights}
\end{figure}

Figure~\ref{fig:loss_weights} shows the effect of varying the weights on reconstruction and decomposition losses on
performance on the Naturally Evaluated / Synthetically Trained task, using the validation set. In all cases, the
KL-divergence loss $\mathcal{L}_{KL}$ was set to a constant (1). The effect of the weight on reconstruction loss is shown in 
Figure~\ref{fig:loss_weights}-(a). When the weight is too small, reconstructions become blurry and the generated digits are not useful
samples for training. When the weight is too large, the network learns to ignore content, reconstructing only from style. In this case,
the generated images look better, but not of the intended class. The blue line in Figure~\ref{fig:loss_weights}-(b) shows the effect of the weight on
predictability minimization, with the reconstruction weight clamped to X. The green dotted line shows performance with no predictability
minimization. When the weight is too small, we hypothesize that predictability minimization can still interfere with the network's ability to
reconstruct, but without providing much benefit in terms of reducing representation redundancy. When the weight is too large, the network
can no longer generate good images. 
Figure~\ref{fig:loss_weights}-(c) shows the effect of both coefficients of $\mathcal{L}_{\LF}$ ($\lambda_1$ and $\lambda_2$) on NEST performance.
Since there are two coefficients, we demonstrate the effect of one when the other is held constant at its best setting.
The left-hand plot shows the effect of $\lambda_1$, which governs the component that filters leakage of style from $z^c$ when generating an image. 
When this parameter is set too low, reconstructions are blurry and not useful for training. We found that as long as this parameter is set to a high
enough value, increasing it further does not appear to affect the quality of generated images. The righthand plot of Figure~\ref{fig:loss_weights}-(c)
shows the effect of $\lambda_2$, which governs the component that filters leakage of content from $z^s$ when generating an image. Again, we find that
as long as this parameter is not set too low, performance is relatively insensitive to its value. This behavior is advantageous for \LF: its
insensitivity to relative weighting means that not much hand-tuning is necessary, while \CC~and \PM~need a great deal of hand tuning.

\subsection{Effect of Predictability Minimization On Content Information in Style}
\begin{figure}
    \centering
    \includegraphics[scale=0.5]{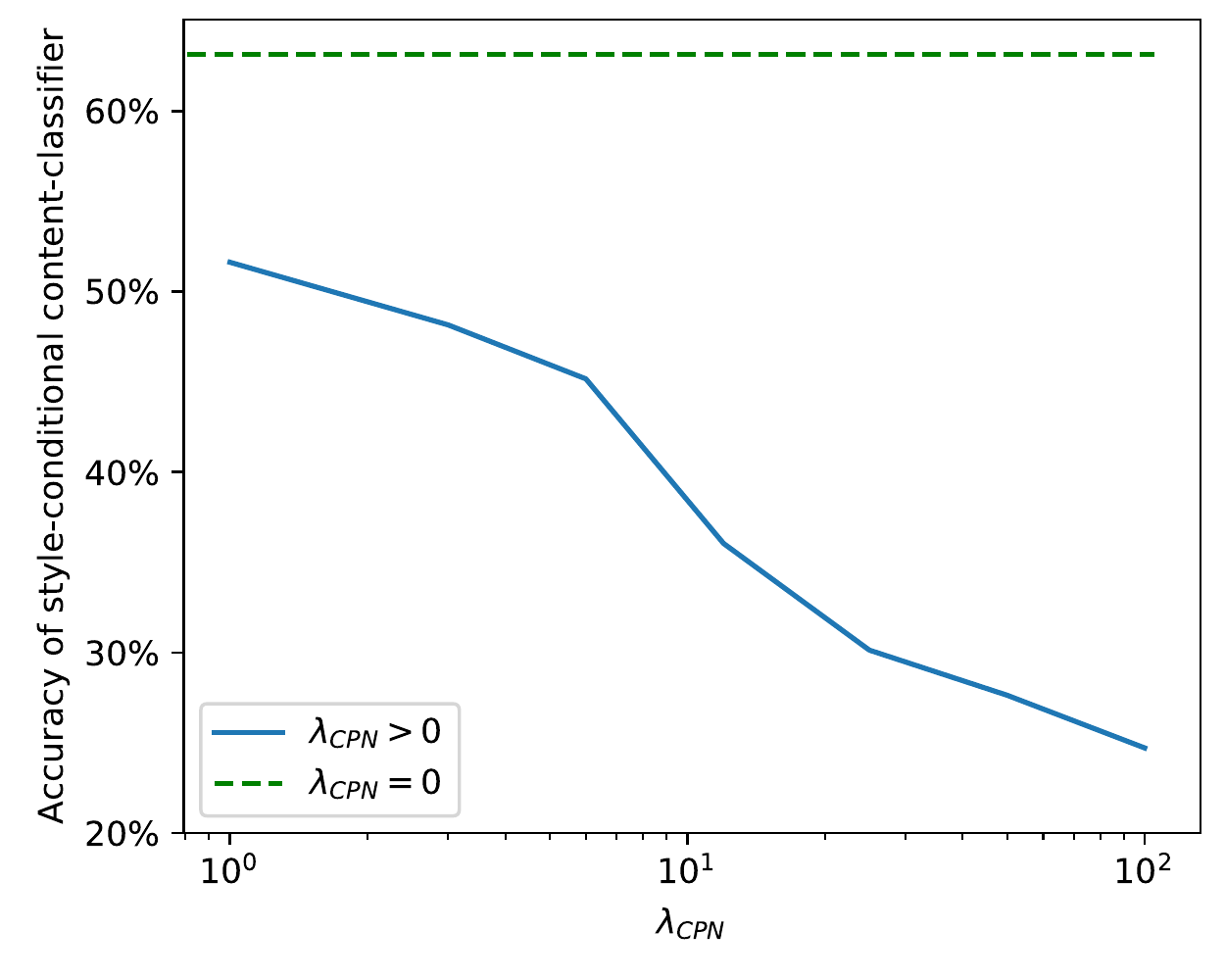}
    \caption{Accuracy of a post-hoc network trained to predict content from style, as a function of weight on Content Prediction Net (CPN)
    objective of \PM.}
    \label{fig:pm_style_conditional_content_classifier}
\end{figure}

If predictability minimization is working properly, it should reduce the amount of content information recoverable from
the style representation. To investigate the effect, we trained several \STOC~models on MNIST using predictability minimization, and
varied the weight on the \PM loss. After training, we train feedforward neural net with one hidden layer that predicts digit-class from
the posterior mean and variance of the style representation. As content information is removed from style, the classifier should be
increasingly inaccurate. The blue line in Figure~\ref{fig:pm_style_conditional_content_classifier} shows style-conditional
content classification accuracy as a function of the weight on the predictability minimization decomposition loss. The dotted green line
shows the accuracy when predictability minimization is disabled. As expected, the accuracy decreases with greater weight on \PM.



\end{document}